# Bayesian Out-Trees

**Tony Jebara**
Columbia University
New York, NY 10027
jebara@cs.columbia.edu

## Abstract

A Bayesian treatment of latent directed graph structure for non-*iid* data is provided where each child datum is sampled with a directed conditional dependence on a single unknown parent datum. The latent graph structure is assumed to lie in the family of directed out-tree graphs which leads to efficient Bayesian inference. The latent likelihood of the data and its gradients are computable in closed form via Tutte's directed matrix tree theorem using determinants and inverses of the out-Laplacian. This novel likelihood subsumes *iid* likelihood, is exchangeable and yields efficient unsupervised and semi-supervised learning algorithms. In addition to handling taxonomy and phylogenetic datasets the out-tree assumption performs surprisingly well as a semi-parametric density estimator on standard *iid* datasets. Experiments with unsupervised and semi-supervised learning are shown on various UCI and taxonomy datasets.

## 1  INTRODUCTION

Many machine learning methods use graph connectivity structure to constrain the dependencies between random variables or between the samples in a dataset. If the graph structure is latent, Bayesian inference or heuristics are used to recover it. This article explores a distribution over a family of directed graphs known as out-trees where Bayesian inference remains efficient. Furthermore, the directed graph connectivity across samples is helpful not only for structured datasets but *iid* datasets as well.

Graph structure is useful to constrain dependencies between random variables (Pearl, 1988; Meila & Jaakkola, 2006), dependencies across samples [1] in a dataset (Roweis & Saul, 2000; Carreira-Perpinan & Zemel, 2004) or even a heterogeneous combination of the two. It may be acceptable to heuristically choose a single graph structure for some problems (Roweis & Saul, 2000) but, in many settings, a Bayesian treatment of latent graph structure can be more precise (Friedman, 1998; Friedman & Koller, 2003; Kemp et al., 2003; Neal, 2003). Tree structures are a particularly efficient family of subgraphs and are relevant in many real-world non-*iid* datasets spanning dynamics, genetics, biology, decision-making, disease transmission and natural language processing (Leitner et al., 1996; Helmbold & Schapire, 1997; Willems et al., 1995; Mau et al., 1999; Koo et al., 2007). Bayesian inference over undirected trees is efficient (Meila & Jaakkola, 2006) however, graph families beyond undirected trees may require approximate inference. For example, recovering an optimal graph is NP-hard for graphs in families with more than 1 parent per node and requires approximation methods like MCMC sampling which may have slow mixing times (Friedman & Koller, 2003).

This article uses graphs primarily to constrain dependency across exchangeable samples in a dataset. We will assume a latent graph structure was responsible for generating the data and assume it lies in the family of directed out-trees. Like undirected trees, this family of directed graphs also benefits from efficient Bayesian inference algorithms. However, the directed aspect of out-trees is not only beneficial for non-*iid* structured datasets like taxonomy trees it also (surprisingly) improves density estimation for *iid* datasets. We conjecture that the directed tree graph structure assumption acts as a flexible semiparametric estimator that overcomes mismatch between a parametric model and an otherwise *iid* dataset.

---

[1]In manifold learning, dependencies across samples in a dataset are often constrained using k-nearest neighbor and/or maximum weight spanning tree subgraphs.

This paper is organized as follows. Section 2 describes how out-trees may emerge sequentially in nature and then presents a computationally convenient generative model for them. Section 3 describes Bayesian inference under the latent out-tree assumption and introduces Tutte's directed matrix tree theorem for efficient inference. Section 4 describes an unsupervised maximum likelihood approach to refining the parameters for the out-tree model. Section 5 describes a semi-supervised approach where the out-tree model can be used to transfer inductive bias from inputs to labels. A variational Bayesian setup for integrating over both structure and parameters is described in Section 6. Experiments with unsupervised learning and semisupervised learning are shown in Section 9 and followed by a brief discussion.

## 2 THE GENERATIVE MODEL

Consider a real-world example of a directed out-tree: the genealogical dataset of neuroscience PhD graduates[2]. This is a population dataset containing $t = 1, \ldots, T$ input samples. Each sample or individual $t$ in this dataset is a node which has a single parent $\pi(t)$, the main doctoral advisor for student $t$. Also, each node $t$ has a corresponding attribute vector $X_t$ associated with it which describes the student (dissertation topic area, year of graduation, institute, etc.). One realistic way such an out tree is generated is in a sequential or temporal manner. A root node labeled $t = 1$ is sampled with a corresponding vector $X_1$ of attributes. Knowing the value $X_1$ for the root, a number of its children nodes are sampled with attributes that depend on the current settings of the parent, $X_1$. These attribute vectors are drawn from a conditional asymmetric distribution $p(X|X_1)$. This is because an advisor is likely to generate students with related dissertation topics, within a finite number of his years of graduation, and so on. This conditional mimics the process of *mutation* in phylogeny. The children then become parents themselves and go on to generate further descendants by sampling again from the conditional distribution $p(X_t|X_{\pi(t)})$. Assume each attribute vector $X_t$ is a 3D Euclidean vector describing the PhD graduate. One choice for the conditional distribution or mutation is the conditioned Gaussian $\mathcal{N}(X_t|\Sigma_{c|\pi}X_{\pi(t)}, I)$ with a fixed correlation parameter $\Sigma_{c|\pi}$ and identity variance. Figure 1 shows some synthetic out-trees with 3D attribute vectors generated by this conditional.

While the above sequential method for generating real-world data is interesting, computational considerations will encourage us to follow a slightly different generative modeling assumption. The data in Figure 1 was actually generated via a non-sequential recipe as follows. We assume an integer $T$ is given that indicates the total number of nodes. Then, from a prior distribution over trees $p(\mathcal{T})$, an undirected tree $\mathcal{T}$ is chosen to connect the $T$ nodes. Then, we choose a root node from the set of nodes $1, \ldots, T$. We then obtain an out-tree by choosing all edges to point away from the root. Next, the attributes of the root are sampled from a marginal distribution $p(X_r)$. Then, traversing from parent to child along the out-tree, we sample the child's attribute vector $X_t$ according to a conditional (mutation) distribution that depends on its parent $X_{\pi(t)}$ denoted by $p(X_t|X_{\pi(t)})$.

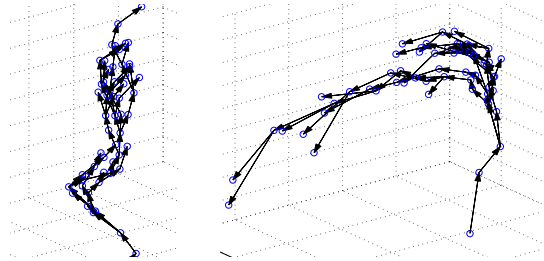

Figure 1: Sampled out-trees using $\mathcal{N}(X_t|\Sigma_{c|\pi}X_{\pi(t)}, I)$.

As is often the case in learning, we assume that some aspects of this generative process are hidden and must be recovered via inference. For instance, we will assume that the tree structure $\mathcal{T}$, the choice of the root and so on are not available to the learner. Instead, as in many real-world datasets, we can only access the node attributes $X_1, \ldots, X_T$ given in some arbitrary ordering. One challenge is to recover the marginal $p(X_r)$ and conditional distributions $p(X_t|X_{\pi(t)})$ from the data. Another challenge is to recover information about the lost connectivity structure $\mathcal{T}$. Another challenge is to recover missing information in some of the attribute vectors, i.e. hidden elements in $X_1, \ldots, X_T$. This article presents efficient Bayesian inference approaches to these problems.

## 3 EFFICIENT DISTRIBUTIONS OVER OUT-TREES

We are given a training dataset containing input samples $X_t$ for $t = 1 \ldots T$ in some arbitrary order. One quantity to evaluate or manipulate is the likelihood of the dataset $p(X_1, \ldots, X_T|\theta, T)$ given some model. A popular method to recover a model of the dataset is to find the model that maximizes the likelihood score. An additional standard assumption most unsupervised methods make is that the dataset is composed of independently identically distributed samples. In other words, $p(X_1, \ldots, X_T|\theta, T) = \prod_{t=1}^{T} p(X_t|\theta)$. This *iid* assumption can often be inappropriate for real datasets. What is a more minimal set of assumptions on the likelihood function we can make? Likelihoods should be non-negative and sum to unity if we integrate over all $X_1, \ldots, X_T$. In addition, since the data arrives in an arbitrary order, a likelihood should be invariant to permutation of the arguments $\{X_1, \ldots, X_T\}$ for any given finite dataset size $T$. This property is called finite exchangeability. It is less strict than infinite exchangeability which is in turn less strict than *iid* sampling. [3] The next section derives a likelihood that satisfies these properties yet generalizes the *iid* setting by assuming data was sampled according to an out-tree data structure. The generative model assumes that we first form a complete tree with $T$ nodes (the number $T$ is known a priori) and then children are sampled from their parents using conditional distributions according to the out-tree.

More formally, define an out-tree as an acyclic graph $\mathcal{T}$ with a set of $T$ vertices $\mathcal{X} = X_1, \ldots, X_T$ and directed edges such that each node $X_t$ has at one parent node $X_{\pi(t)}$ and the root has no parents. Note, here we abuse notation and take $X_t$ to refer to the node corresponding to the $t$'th sample as well as its attribute vector interchangeably. Rooted out-trees are trees with directed edges pointing away from a well-defined root. For instance, $X_1 \leftarrow X_2 \leftarrow X_3$ is an out-tree rooted at $X_3$. Conversely, rooted in-trees have all directed edges from other nodes point towards the root. The previous 3-chain example is thus also an in-tree rooted at node $X_1$. Many directed trees are neither in-trees nor out-trees. For instance, the tree $X_1 \rightarrow X_2 \leftarrow X_3 \rightarrow X_4$ is a valid directed tree but neither a rooted in-tree nor a rooted out-tree. For each choice of a root, the set of rooted out-trees forms a disjoint set of $T^{T-2}$ directed trees. Therefore, there are $T^{T-1}$ out-trees for $T$ nodes.

If we knew the latent out-tree structure $\mathcal{T}$ that generated our $T$ samples, the likelihood of the data under the generative assumptions of Section 2 would factorize as a product of conditionals of each node given its parent. However, in general, the structure is unknown. Consider treating structure as a random variable and using Bayes' rule to obtain a posterior distribution over out-trees as follows:

$$p(\mathcal{T}|\mathcal{X}) = \frac{p(\mathcal{X}|\mathcal{T})p(\mathcal{T})}{p(\mathcal{X})} = \frac{\prod_{t=1}^{T} p(X_t|X_{\pi(t)})p(\mathcal{T})}{p(X_1, \ldots, X_T)}.$$

A typical assumption is that the prior over out-tree structures is chosen to be uniform yielding $p(\mathcal{T}) =$

$\frac{1}{\text{card}(\mathcal{T})} = \frac{1}{T^{T-1}}$. This is merely a normalized constant distribution over all possible out-trees. We rewrite the posterior over out-trees as follows:

$$p(\mathcal{T}|\mathcal{X}) = \frac{p(\mathcal{X}|\mathcal{T})}{p(\mathcal{X})T^{T-1}} = \frac{1}{Z} \prod_{t=1}^{T} p(X_t|X_{\pi(t)}). \quad (1)$$

where we have defined the *partition function $Z$* that ensures that the likelihood term sums to unity over all possible out-trees:

$$Z = p(\mathcal{X})T^{T-1} = \sum_{\mathcal{T} \in \Gamma} \prod_{t=1}^{T} p(X_t|X_{\pi(t)}).$$

Here, $\mathcal{T}$ enumerates over the set of all out-trees, $\Gamma$. This is an unwieldy computation since there are $T^{T-1}$ possible out-trees connecting $T$ observation vertices. Instead, we consider breaking up the summation into all possible choices of the root of the out-tree $r = 1 \ldots T$ and a summation over the subset $\Gamma_r$ of all $T^{T-2}$ out-trees rooted at node $r$. It is straightforward to show that all subsets of out-trees with different roots are distinct, in other words $\Gamma_i \cap \Gamma_j = \{\}$ if $i \neq j$. Furthermore, their union forms the set of all out-trees $\Gamma = \cup_{j=1}^{T} \Gamma_j$. Thus, the partition function $Z$ is given by the following sum:

$$Z = \sum_{r=1}^{T} \sum_{\mathcal{T}_r \in \Gamma_r} \prod_{t=1}^{T} p(X_t|X_{\pi(t)})$$
$$= \sum_{r=1}^{T} p(X_r) \sum_{\mathcal{T}_r \in \Gamma_r} \prod_{t \neq r}^{T} p(X_t|X_{\pi(t)})$$

where we have used the property that the root has no parent node. To efficiently recover $Z$ we will instead recover the individual components of the above sum over $r$:

$$Z_r = \sum_{\mathcal{T}_r \in \Gamma_r} \prod_{t \neq r}^{T} p(X_t|X_{\pi(t)})$$

by making an appeal to the directed variant of Kirchoff's *Matrix Tree Theorem*, namely Tutte's *Directed Matrix Tree Theorem* (West, 1996). The directed matrix tree theorem does not quite sum over all directed trees. It sums over a *subset*: rooted out-trees. To apply Tutte's theorem we compute an asymmetric $\beta$ weight matrix of size $T \times T$ populated by all pairwise conditional probabilities $\beta_{uv} = p(X_u|X_v)$. Note that we will assume $\beta_{vv} = 0$ since there are no edges between a node and itself. The matrix $\beta$ allows us to rewrite $Z_r$ as a product of edges in the tree instead of a product of nodes:

$$Z_r = \sum_{\mathcal{T}_r \in \Gamma_r} \prod_{uv \in \mathcal{T}_r} \beta_{uv}.$$

The out-tree Laplacian matrix $Q$ is then obtained as follows:

$$Q = diag(\beta \vec{1}) - \beta.$$

Here, take $\vec{1}$ to be the ones column vector and note that the $diag(\vec{v})$ operator gives a diagonal matrix with $\vec{v}$ on its diagonal. Note that this *out Laplacian* is not symmetric since $\beta$ is not symmetric. Similarly, the *in Laplacian* is given by $Q_{in} = diag(\vec{1}\beta) - \beta$. The directed matrix tree theorem asserts that the number (or weight) of out-trees rooted at node $r$ is $Z_r$ and is given by the matrix cofactor $[Q]_r$ obtained by deleting the $r$'th row and $r$'th column of the matrix $Q$. The precise formula is:

$$Z_r = |[Q]_r| = |[diag(\beta \mathbf{1}) - \beta]_r|.$$

Reinserting this formula into the above gives the total partition function as:

$$Z = \sum_{r=1}^{T} p(X_r) Z_r = \sum_{r=1}^{T} p(X_r) |[diag(\beta \mathbf{1}) - \beta]_r|$$

which is now efficient to evaluate. Interestingly, $Z$ is the sum of determinants of the minors of the Laplacian. If $\beta$ is symmetric, all terms in the summation above are identical and we need to only work with a single determinant of the $T \times T$ matrix. A symmetric $\beta$ would emerge, for example, if we chose symmetric conditional distributions $p(X_u|X_v) = p(X_v|X_u)$. In addition, it is known that the log determinant of a symmetric Laplacian matrix is a concave function of the edge-weights (Jakobson & Rivin, 2002). In the asymmetric case, however, the log-partition function does not preserve concavity.

A naive implementation recovers $Z$ in $\mathcal{O}(T^4)$ however it is possible in cubic time as in (Jebara & Long, 2005; Koo et al., 2007) via straightforward applications of Woodbury's formula. This is done by creating a matrix of size $(T + 1) \times (T + 1)$ called $\hat{Q}$:

$$\hat{Q} = \begin{bmatrix} 1 & \vec{p}^T \\ -\vec{p} & Q \end{bmatrix}$$

where $\vec{p}$ is a unit-normalized vector of length $T$ whose entries are proportional to the root probabilities:

$$\vec{p}(r) = \frac{p(X_r)}{\sum_{r=1}^{T} p(X_r)}.$$

The partition function can then be computed by a single evaluation of the matrix determinant as $Z = (\sum_{r=1}^{T} p(X_r)) |\hat{Q}|$ which is $\mathcal{O}((T + 1)^3)$ although faster methods are also possible (Kaltofen & Villard, 2004). This is an improvement over the summation of smaller determinants which required $\mathcal{O}(T^4)$. In addition, for numerical reasons, we use the logarithm of the partition function. This is recovered via the trace of the matrix logarithm (or the sum of the log-singular values after an SVD) as:

$$\ln Z = \ln(\sum_{r=1}^{T} p(X_r)) + tr\left(\ln \begin{bmatrix} 1 & \vec{p}^T \\ -\vec{p} & Q \end{bmatrix}\right) \quad (2)$$

which takes $\mathcal{O}((T + 1)^3)$ time. This computation is more efficient than enumerating over all $T^{T-1}$ out-trees as in the normalized posterior of Equation 1 and makes it possible to consider datasets beyond a thousand points. To scale further, a wide set of approximate methods for large matrices can be leveraged including Nystrom methods (Williams & Seeger, 2001; Drineas & Mahoney, 2005) and column sampling. These methods will be investigated in future work but were not necessary for initial experiments.

## 4 MAXIMUM *tdid* LIKELIHOOD

An interesting property of the partition function $Z$ is that it forms a finitely exchangeable *tdid* or tree dependent identically distributed likelihood. The likelihood of the data is $p(\mathcal{X}) = ZT^{1-T}$ or more explicitly:

$$p(X_1, \ldots, X_T) = \frac{1}{T^{T-1}} \sum_{r=1}^{T} p(X_r) |[diag(\beta \mathbf{1}) - \beta]_r| \quad (3)$$

which degenerates into the *iid* likelihood when the conditional dependence between parent and child nodes is extinguished.

**Theorem 1** *If the conditional dependence of a child node given a parent node degenerates into the marginal $p(X_t|X_{\pi(t)}) \to p(X_t)$ the* tdid *likelihood simplifies into the* iid *likelihood.*

**Proof 1** *Work backwards by writing the* tdid *likelihood in terms of a product over nodes:*

$$p(X_1, \ldots, X_T) = \frac{1}{T^{T-1}} \sum_{r=1}^{T} p(X_r) \sum_{\mathcal{T}_r \in \Gamma_r} \prod_{t \neq r}^{T} p(X_t|X_{\pi(t)}).$$

*Removing the dependence on the parent produces:*

$$p(X_1, \ldots, X_T) = \frac{1}{T^{T-1}} \sum_{r=1}^{T} \sum_{\mathcal{T}_r \in \Gamma_r} \prod_{t=1}^{T} p(X_t)$$

*which then simplifies into the* iid *likelihood:*

$$p(X_1, \ldots, X_T) = \frac{\sum_{r=1}^{T} T^{T-2}}{T^{T-1}} \prod_{t=1}^{T} p(X_t) = \prod_{t=1}^{T} p(X_t).$$

Thus a generalization of *iid* likelihood emerges by integrating over a latent out-tree sampling structure. One natural way of performing unsupervised learning is to maximize this *tdid* likelihood to recover, for instance, a good setting of the parameters $\theta$ that govern the conditional distribution of child given parent. Equation 3 acts as a novel maximum likelihood estimator. We rewrite the *tdid* likelihood to make the dependence on the conditional distribution's parameters $\theta$ more explicit:

$$p(X_1, \ldots, X_T | \theta) = \frac{Z(\theta)}{T^{T-1}}.$$

This likelihood satisfies certain desiderata outlined earlier. First, it is invariant to reordering of $\{X_1, \ldots, X_T\}$ and therefore is finitely exchangeable. Furthermore, it is easy to verify that $p(\mathcal{X}) \geq 0$ and sums to unity when integrated over all possible $X_1, \ldots, X_T$.

We next consider maximum likelihood unsupervised learning where we find a $\theta$ that produces a large $p(\mathcal{X} | \theta)$. We maximize the log *tdid* likelihood using gradient ascent on the parameters. The gradient for any scalar parameter $\theta_i$ is given by the chain rule applied to Equation 2:

$$\frac{\partial \ln Z}{\partial \theta_i} = \frac{1}{\sum_{r=1}^{T} p(X_r)} \sum_{r=1}^{T} \frac{\partial p(X_r)}{\partial \theta_i} + \mathrm{tr}\left(\hat{Q}^{-1} \frac{\partial \hat{Q}}{\partial \theta_i}\right)$$

The main computational requirement for evaluating the gradient is the $\mathcal{O}((T+1)^3)$ matrix inversion. However, the computation of the gradient can easily be approximated for further efficiency.

Given an initial guess for $\theta$, it is possible to follow the gradient or perform line-search. Line search is convenient since evaluating determinants is faster than matrix inversion. Note that maximum likelihood with incomplete information is being performed since we never require anything more than the $\{X_1, \ldots, X_T\}$ population data (there is no additional information about the tree structure).

While any exponential family distribution could be used to specify the marginal and conditional distributions, we focus on the Gaussian case. The following marginal-conditional decomposition of its parameters holds $\theta = \{\mu_c, \mu_\pi, \Sigma_{c|\pi}, \Sigma_{cc}, \Sigma_{\pi\pi}\}$. These are two vectors in $\mathbb{R}^D$ and three matrices in $\mathbb{R}^{D \times D}$. We further assume matrices $\Sigma_{cc}$ and $\Sigma_{\pi\pi}$ are positive definite. This gives the following Gaussian probabilities for the root nodes:

$$p(X_r | \theta) = \mathcal{N}(X_r | \mu_\pi, \Sigma_{\pi\pi}),$$

and the following conditional Gaussian probability of child given parent:

$$p(X_t | X_{\pi(t)}, \theta) = \mathcal{N}(X_t | \Sigma_{c|\pi} X_{\pi(t)} + \mu_c, \Sigma_{cc}).$$

If we are given a parameter setting and if all $X_t$ variables are observed, it is straightforward to apply these formulae. The resulting probabilities are inserted into the out Laplacian which efficiently recovers the likelihood value or the gradients for unsupervised learning.

Given the gradient and the likelihood evaluation, we can now readily maximize the *tdid* likelihood. However since it is not concave in the exponential family case except when *tdid* degenerates into *iid*, we prefer to initialize with the *iid* solution. For example, in the Gaussian case, a reasonable initialization for $\theta$ is to learn the model under *iid* assumptions for the seed model and then perform maximum *tdid* likelihood thereafter.

Once we have learned a model $\theta^*$ from training data $\mathcal{X}$, we evaluate the test likelihood on new data $\tilde{\mathcal{X}} = \{\tilde{X}_1, \ldots, \tilde{X}_U\}$ according to:

$$p(\tilde{\mathcal{X}} | \mathcal{X}, \theta^*) = \frac{p(\tilde{\mathcal{X}}, \mathcal{X} | \theta^*)}{p(\mathcal{X} | \theta^*)} = \frac{Z_{\mathcal{X} \cup \tilde{\mathcal{X}}}(\theta^*)}{Z_{\mathcal{X}}(\theta^*)} \frac{T^{(T-1)}}{(T+U)^{(T+U-1)}}.$$

It is straightforward to show that this quantity still integrates to one when we integrate over $\tilde{\mathcal{X}}$. This simply involves computing the partition function for the test data aggregated with the training data $Z_{\mathcal{X} \cup \tilde{\mathcal{X}}}$ relative to the partition function for the training data alone $Z_{\mathcal{X}}$. Both these quantities involve the determinant formula we outlined. Contrast the above test likelihood score to the traditional test likelihood score produced by an *iid* model which simplifies due to factoring:

$$p_{iid}(\tilde{\mathcal{X}} | \mathcal{X}, \theta^*) = \frac{\prod_{j=1}^{T} p(X_j | \theta^*) \prod_{i=1}^{U} p(\tilde{X}_i | \theta^*)}{\prod_{j=1}^{T} p(X_j | \theta^*)}.$$

In *iid* each test point data is independent of the training data and all other test points given the model parameters $\theta^*$. Thus, the *tdid* likelihood is semi-parametric since, in addition to depending on parameters $\theta$, there is a non-parametric dependence on other training and test points. In fact, if we set the conditional Gaussians such that $\Sigma_{c|\pi} = I$ and $\mu_{cc} = 0$, *tdid* can mimic non-parametric Parzen estimation.

## 5 SEMI-SUPERVISED OUT-TREES

Another application of the latent out-tree assumption is in semi-supervised learning problems (Kemp et al., 2003) where output labels are given in addition to input samples. Consider a label $y_t$ which is generated from a mutation process over the branches of the tree $\mathcal{T}$ just as the attributes of the node $X_t$ are also mutations from their parents. This mutation process defines a distribution over possible labels. For instance, $y_t$ may indicate if an individual in a population has diabetes and $X_t$ is a vector of anatomical features for the individual. Instead of generating a label that depends only on the input, we could also consider dependence of a child label on a parent input and a parent label. In other words, for training samples $X_t$ and corresponding labels $y_t$ for $t = 1 \ldots T$, we have the following likelihood for a known out-tree $\mathcal{T}$:

$$p(X_1, y_1, \ldots, X_T, y_T | \mathcal{T}) = \prod_{t=1}^{T} p(X_t, y_t | X_{\pi(t)}, y_{\pi(t)}). \quad (4)$$

The derivations for the partition function proceed as before but now involve a directed Laplacian built from edge weights using conditionals between both inputs and outputs, i.e. $\beta_{uv} = p(X_u, y_u | X_v, y_v)$. We may make more restrictive factorization assumptions on these conditional relationships, for instance, $y_t$ might depend *only* on $X_t$. A more interesting assumption is $y_t$ is independent of input data altogether and is only conditionally dependent on its parent's label $y_{\pi(t)}$. In other words, the mutation conditional simplifies into $p(y_t | y_{\pi(t)}) p(X_t | X_{\pi(t)})$. In an *iid* classification setting, this radical assumption would make learning impossible since output data is independent from input data. However, in a latent out-tree setting, inputs and outputs are only conditionally independent *given* tree structure. When tree structure is unknown, dependence between inputs and outputs emerges without an explicit relationship between the input and output spaces (parametric or otherwise). This is because $X$ and $y$ are sampled given the parent random variable $\mathcal{T}$, i.e. $X \leftarrow \mathcal{T} \rightarrow y$. Therefore, observing the input induces a posterior on $\mathcal{T}$ which subsequently induces a posterior on labels. This is particularly useful when we cannot make explicit assumptions about the parametric relationship between the input and output spaces. In these settings, the latent out-tree may be a good source of inductive bias to couple inputs to outputs especially if the number of labeled outputs is small.

To recover the settings of the unobserved $y_t$ labels, one approach is to maximize likelihood while integrating over $\mathcal{T}$. Assume we have observed the labels for the training points $\mathcal{Y} = \{y_1, \ldots, y_T\}$ but not for the test points $\tilde{\mathcal{Y}} = \{\tilde{y}_1, \ldots, \tilde{y}_U\}$. To predict labels, we need the conditional posterior over unobserved labels given all other observed data, $p(\tilde{\mathcal{Y}} | \mathcal{X}, \tilde{\mathcal{X}}, \mathcal{Y}, \theta)$ as follows:

$$p(\tilde{\mathcal{Y}} | \mathcal{X}, \tilde{\mathcal{X}}, \mathcal{Y}, \theta) = \frac{p(\mathcal{X}, \tilde{\mathcal{X}}, \mathcal{Y}, \tilde{\mathcal{Y}} | \theta)}{\sum_{\tilde{\mathcal{Y}}} p(\mathcal{X}, \tilde{\mathcal{X}}, \mathcal{Y}, \tilde{\mathcal{Y}} | \theta)}.$$

Instead of maximizing the above conditional likelihood, we simply maximize the joint latent likelihood (a common simplification in many learning frameworks) to recover parameters $\theta$ and unknown labels:

$$p(\mathcal{X}, \tilde{\mathcal{X}}, \mathcal{Y}, \tilde{\mathcal{Y}} | \theta) = \frac{Z(\tilde{\mathcal{Y}}, \theta)}{(T + U)^{(T+U-1)}}$$

where we have rewritten the partition function in terms of both the unknown $\mathcal{Y}$ and unknown $\theta$ which need to be specified to construct the out Laplacian $\hat{Q}$. We initialize both randomly and then maximize the partition function using gradient ascent for $\theta$ as in the unsupervised learning case and maximize over unknown labels $\tilde{\mathcal{Y}}$ by greedily flipping individual labels to increase the partition function. This iterative hill climbing scheme produces a final set of parameters and labels. While investigating a label flip in the output, it is useful to avoid full matrix inversion and full matrix determinants since each label flip only involves a rank 1 change to the out Laplacian matrix $\hat{Q}$ and thus each step of hill climbing requires no more than quadratic time. This and intermediate caching of results allows efficient prediction of labels.

## 6 VARIATIONAL BAYES

So far we have considered the latent *tdid* likelihood which involves agnostically integrating over all structures $\mathcal{T}$. However, we have assumed that the parameters $\theta$ are given or are recovered by a point estimator such as maximum likelihood. A more thorough Bayesian approach is to consider integrating over *both* parameters $\theta$ and structure $\mathcal{T}$ after introducing a prior distribution on both. In such a setting, the nonparametric density estimator becomes reminiscent of other nonparametric Bayesian methods such as Dirichlet processes (Teh et al., 2004; Neal, 2003; Ferguson, 1973) and infinite mixture models (Rasmussen, 1999; Beal et al., 2002). The joint integration over parameters and structure recovers the evidence of the data $p(\mathcal{X})$ or equivalently the log-evidence $\mathcal{E} = \ln p(\mathcal{X})$ (Friedman & Koller, 2003). Consider first splitting the parameters $\theta$ into those adjusting the distribution over root $\theta_m$ and those adjusting the distribution of child given parent $\theta_c$. We assume the root $p(X_r | \theta_m)$ and conditional distributions $p(X_t | X_{\pi(t)}, \theta_c)$ are in the exponential family and the priors on their parameters $p(\theta) = p(\theta_m) p_c(\theta_c)$ are conjugate. Integrating with a uniform structure prior $p(\mathcal{T}) = \frac{1}{T^{T-1}}$ and making the natural assumption that the prior over structure and parameters factorizes yields:

$$\mathcal{E} = \ln \int_{\theta} \sum_{r=1}^{T} p(X_r | \theta_m) \sum_{\mathcal{T}_r} \prod_{t \neq r}^{T} p(X_t | X_{\pi(t)}, \theta_c) p(\mathcal{T}) p(\theta)$$

which unfortunately is an intractable quantity. We instead consider a lower bound on the evidence. This is done by introducing variational distributions, for instance, the distribution $q(r)$ over choices for the root and applying Jensen.

$$
\begin{aligned}
\mathcal{E} \geq \sum_{r=1}^{T} q(r) \ln \int_{\theta} p(X_r|\theta_m) \sum_{\mathcal{T}_r}^{T} \prod_{t \neq r}^{T} p(X_t|X_{\pi(t)}, \theta_c) p(\theta) \\
+ H(q) - (T-1) \ln T
\end{aligned}
$$

We also introduce variational distributions over $r = 1, \ldots, T$ out-trees each rooted at node $r$ which we denote by $q_r(\mathcal{T}_r)$ and a variational distribution over the parameters $q_c(\theta_c)$. Re-applying Jensen's inequality produces:

$$
\begin{aligned}
\mathcal{E} \geq & \sum_{r} q(r) \ln \int_{\theta_m} p(X_r|\theta_m) p(\theta_m) - (T-1) \ln T \\
& + \sum_{r, \mathcal{T}_r} q(r) q_r(\mathcal{T}_r) \sum_{t \neq r}^{T} \int_{\theta_c} q_c(\theta_c) \ln p(X_t|X_{\pi(t)}, \theta_c) \\
& + H(q) + \sum_{r} q(r) H(q_r) - KL(q_c \| p_c).
\end{aligned}
$$

Above, $H$ denotes the Shannon entropy and $KL$ denotes the Kullback-Leibler divergence. Update rules for each variational distribution iteratively maximize the lower bound by taking derivatives and setting to zero. We update the density over out-trees rooted at node $r$ via:

$$
q_r(\mathcal{T}_r) = \frac{1}{Z_r} \prod_{t \neq r}^{T} e^{\int_{\theta_c} q_c(\theta_c) \ln p(X_t|X_{\pi(t)}, \theta_c)}.
$$

As in Section 3 this can be rewritten as a product over edges in the out-tree $\mathcal{T}_r$ and summarized simply by a $T \times T$ matrix $\beta$ whose off diagonal entries are $\beta_{uv} = \int_{\theta_c} q_c(\theta_c) \ln p(X_u|X_v, \theta_c)$. For exponential family $p(X_u|X_v, \theta_c)$, such integrals are easy to solve (Box & Tiao, 1992). Each $Z_r$ is also straightforward to recover using Tutte's theorem. The update for the $q(r)$ distribution is:

$$
\begin{aligned}
q(r) \propto & \ e^{H(q_r)} \int_{\theta_m} p(X_r|\theta_m) p(\theta_m) \\
& \times e^{\sum_{\mathcal{T}_r} q_r(\mathcal{T}_r) \sum_{t \neq r}^{T} \int_{\theta_c} q_c(\theta_c) \ln p(X_t|X_{\pi(t)}, \theta_c)}
\end{aligned}
$$

where the entropy $H(q_r)$ and the expectation over $q_r(\mathcal{T}_r)$ are efficient to compute from the $\beta$ matrix (Meila & Jaakkola, 2006). Furthermore, the integrals $\int_{\theta_m} p(X_r|\theta_m) p(\theta_m)$ are known for exponential families. We update the distribution over parameters via:

$$
\begin{aligned}
q_c(\theta_c) & \propto \ p(\theta_c) e^{\sum_{r, \mathcal{T}_r} q(r) q_r(\mathcal{T}_r) \sum_{t \neq r}^{T} \ln p(X_t|X_{\pi(t)}, \theta_c)} \\
& = \ p(\theta_c) \prod_{u \neq v} p(X_u|X_v, \theta_c)^{\sum_{r, \mathcal{T}_r} q(r) q_r(\mathcal{T}_r) \delta(uv \in \mathcal{T}_r)}.
\end{aligned}
$$

This is simply the prior times a product over all pairs of data-points likelihoods with different weights

$q_c(\theta_c) \propto p(\theta_c) \prod_{u \neq v} p(X_u|X_v, \theta_c)^{W_{uv}}$. These weights are recovered easily from the current $\beta$ matrix.

A variational Bayesian treatment is possible over joint out-tree structure and parameters. The method allows us to refine a lower bound on evidence and permits full (nonparametric) Bayesian estimation with out-trees.

# 7 STATIONARY MUTATION

It is helpful to distinguish some important differences between the Bayesian averaging over out-trees here and the Bayesian inference of tree belief networks presented in (Jaakkola et al., 1999; Meila & Jaakkola, 2006). While elegantly providing a tractable computation of the Bayesian inference, (Meila & Jaakkola, 2006) makes no requirement on the stationarity of the conditional distribution $p(X_t|X_{\pi(t)}, \theta_c)$ which is a key distinguishing component of the out-tree framework. In other words, in (Jaakkola et al., 1999; Meila & Jaakkola, 2006) each conditional has its own parameter $\theta_t$ and the samples are drawn from custom conditionals $p(X_t|X_{\pi(t)}, \theta_t)$. If one assumes *decomposable* priors, the Bayesian evidence and Bayesian inference jointly over parameter and structure is elegantly tractable. However, it explores distinct $\theta_t$ for all conditionals. This article introduces the constraint that $\theta_t = \theta_{t'} = \theta_c$ for all $t = 1, \ldots, T$ except for the root node. This constraint greatly restricts the model and assumes that the mutation distribution is stationary across all samples. In other words the parameters of the conditional are fixed. This is a key difference and permits us to recover the *iid* setting as a special case when the conditional dependence is extinguished. The unrestricted Bayesian inference method in (Meila & Jaakkola, 2006) can be seen as a step in the variational Bayesian procedure since the update rule for the distribution $q_c(\theta_c)$ collapses the individual conditionals into a single $\theta_c$ model.

# 8 HILBERT GAUSSIANS

In addition to the Gaussian, many exponential family choices for the marginal distribution over root attributes $p(X_r|\theta_m)$ and for the conditional $p(X_u|X_v, \theta_c)$ are possible and computationally convenient. One useful feature of the Gaussian is that it is readily converted into conditional form and leads to a flexible linear relationship between parent and child which is determined primarily by the variable $\Sigma_{c|\pi}$. To go beyond this linear relationship, we may use a mapping on the features or attributes of the parent node which in no way changes the normalization properties of the Gaussian. Thus, we may consider first mapping the parent's features into a higher dimensional vector rep-

| Dataset $(D,T)$ | Spiral (3,534) | Heart (13,139) | Diabetes (8,268) | Liver (6,200) |
|---|---|---|---|---|
| Parzen | -5.61e3 | -1.94e3 | -6.25e3 | -3.41e3 |
| GMM-1 | -1.36e3 | -2.02e4 | -2.12e5 | -2.53e4 |
| GMM-2 | -1.36e3 | -3.23e4 | -2.85e5 | -1.88e4 |
| GMM-3 | -1.19e3 | -2.50e4 | -4.48e5 | -2.79e4 |
| GMM-4 | -7.98e2 | -1.68e4 | -2.03e5 | -2.62e4 |
| GMM-5 | -6.48e2 | -3.15e4 | -3.40e5 | -3.23e4 |
| GMM-$\infty$ | -4.86e2 | -4.02e2 | -8.22e2 | -4.56e2 |
| *tdid* | -3.91e2 | -5.29e2 | -8.87e2 | -4.99e2 |

Table 1: Gaussian test log-likelihoods using RBF Parzen estimators, EM mixtures of Gaussians, the $\infty$ Gaussian mixture model, and the *tdid* estimator.

resentation $\phi(X_{\pi(t)})$ of dimensionality $\mathbb{R}^H$ and then learning a matrix $\Sigma_{c|\pi}$ of size $D \times H$. This is a generalized linear conditional model that can capture more complex relationships between the parent and child nodes. In such a setting, the conditional Gaussian relationship need not be represented explicitly in the space of $\phi(X_{\pi(t)})$ but only implicitly in kernelized form over each dimension of the input space:

$$p(X|X_\pi) = \prod_{d=1}^{D} \mathcal{N}\left(X(d) \left| \sum_{t=1}^{T} \alpha_{t,d} k(X_\pi, X_t) + \mu_d, \sigma_d \right.\right)$$

where the unconditional means $\mu_d$, variances $\sigma_d$ and weights $\alpha_{t,d}$ are scalars for $t = 1, \ldots, T$ and $d = 1, \ldots, D$, the latter of which indexes the dimensions of the attributes. Furthermore, the function $k(.,.)$ can be any kernel that maps a pair of inputs in the sample space into a scalar measurement of affinity. This gives a general way of extending the linearity assumptions in the conditional model. Instead of making the conditional dependence linear in the values of the parent attributes, we can explore linearity in any *features* of the parent attributes which leads to another source of nonparametric flexibility in the estimator.

## 9 EXPERIMENTS

To visualize the out-tree model's ability to fit data, we estimated marginal and conditional Gaussian parameters using the latent likelihood for the UCI Spiral dataset in Figure 2(a). Once the $\theta = \{\mu_c, \mu_\pi, \Sigma_{c|\pi}, \Sigma_{cc}, \Sigma_{\pi\pi}\}$ parameters were recovered (a total of 27 scalar degrees of freedom), they were used to generate synthetic datasets of 600 samples in Figures 2(b), (c) and (d). In the last example, the matrix $\Sigma_{cc}$ was reduced to sample a spiral with less noise. Notice how the datasets can produce slightly different spirals that may have more or fewer turns but still maintain the appropriate overall shape.

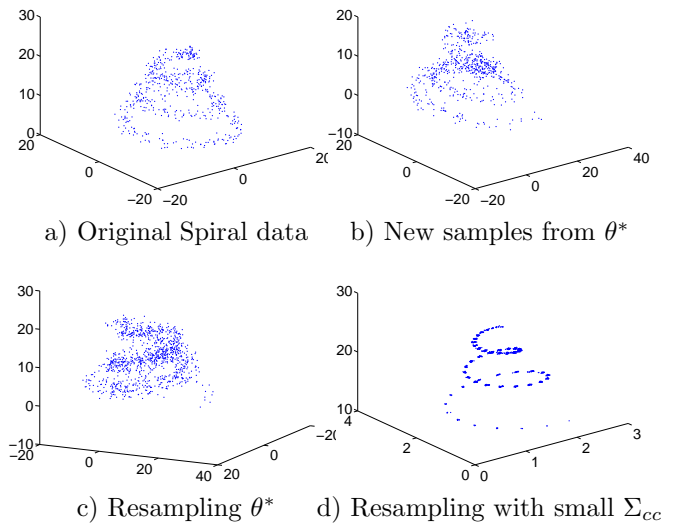

a) Original Spiral data    b) New samples from $\theta^*$

c) Resampling $\theta^*$    d) Resampling with small $\Sigma_{cc}$

Figure 2: Spiral data density estimation.

We next show more quantitative unsupervised density estimation experiments on standard UCI datasets where a large test log-likelihood implies a better density estimate. These experiments closely follow the format in (Jebara et al., 2007). Table 1 summarizes the results with various Gaussian models including the marginal-conditional Gaussian model for the *tdid* out-tree approach. On 4 standard datasets (we only use one class for labeled datasets), the test log-likelihood was evaluated after using a variety of density estimators. These estimators include a nonparametric Parzen RBF estimator with a varying scale parameter $\sigma$. In addition, a mixture of 1 to 5 Gaussians were fit using Expectation Maximization to maximize *iid* likelihood. Comparisons are also shown with semiparametric density estimators like the infinite mixture of Gaussians (Rasmussen, 1999). Cross-validation was used to choose the $\sigma$, and EM local minimum (from ten initializations), for the Parzen and EM approaches respectively. Similarly, cross-validation was used to early-stop the Bayesian out-tree maximum likelihood gradient ascent procedure although this did not have a large effect on performance. Train, cross-validation and test split sizes where 80%, 10% and 10% respectively. The 10 fold averaged test log-likelihoods show that the new method outperforms traditional mixture modeling and Parzen estimators and is comparable to semiparametric methods such as the infinite Gaussian mixture ($iid - \infty$) model (Rasmussen, 1999). Despite the cubic time linear algebra steps for *tdid* estimation, the infinite Gaussian mixture model was the most computationally demanding method.

In a semi-supervised learning problem, we evaluated how well the latent out-tree structure works for clas-sification and its ability to perform inductive transfer. First, $\theta$ is learned from only input samples in an unsupervised manner and the Gaussian parameters $\Sigma_{c|\pi}$ and $\Sigma_{cc} \propto I$ are recovered as above by maximizing the partition function. Then, observed labels are used in the following conditionals to construct out Laplacian:

$$p(X_u|X_v) = \mathcal{N}(X_u|\Sigma_{c|\pi}X_v, \sigma I)$$
$$p(y_u|y_v) = \alpha I(y_u = y_v) + (1-\alpha)I(y_u \neq y_v).$$

Here, the mutation on the outputs is independent of the mutation on the inputs and is simply built from indicator functions with a parameter $\alpha$ which controls the *stickiness* of the label across parent to child. Since only some labels are known, the unknown ones are initialized randomly and improved by maximizing $Z$. This is done with $\Sigma_{c|\pi}$ and $\Sigma_{cc} \propto I$ locked from their unsupervised values. The unknown discrete labels are greedily explored to further increase the partition function. For comparison, support vector machines were trained on the labeled $X$ and $y$ data.

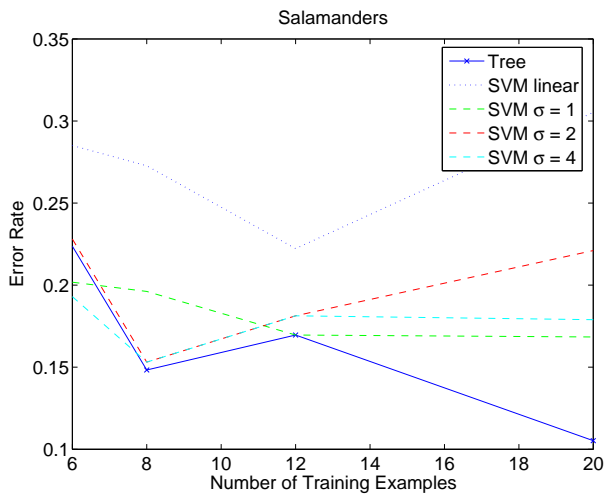

Figure 3: Labeling error rates (averaged over tasks) for Out-Trees and SVMs for salamanders taxonomy.

To evaluate the semi-supervised learning method, two taxonomic datasets (Salamanders and Crustaceans) were used as in (Kemp et al., 2003). These are groups of $T = 30$ and $T = 56$ species nodes and $D = 19$ and $D = 74$ attribute dimensions respectively. Each has a number of discrete attributes describing the external anatomy of each species. These datasets do not have class labels. Therefore each attribute was in turn used as a label to be predicted from the input data. Each dataset therefore generates $D$ discrete prediction tasks. To reduce dimensionality and avoid redundancies (some attributes have the same settings as the target predictions), the remaining $D - 1$ attributes were converted into 3D coordinates using PCA before being used as inputs (both the SVM and the out-tree method are similarly hindered by the resulting loss of information). The input attributes for each problem is a set of 3D vectors $X_1, \ldots, X_T$ and the targets are the original discrete-valued $y_1, \ldots, y_T$ labels.

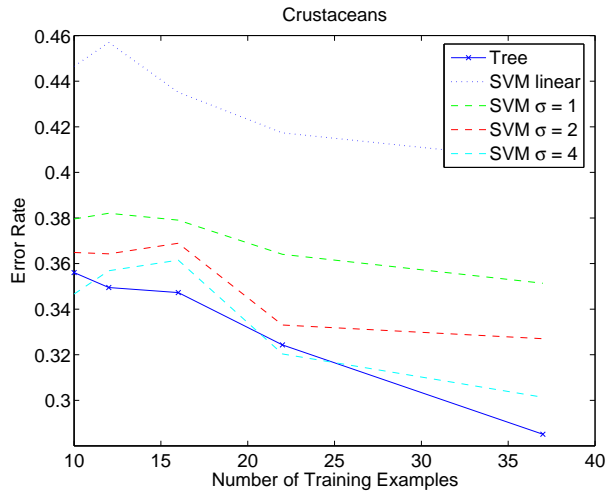

Figure 4: Labeling error rates (averaged over tasks) for Out-Trees and SVMs for crustaceans taxonomy.

Each task was split into training and testing components and the out-tree model was fit and used to find labels. Results were compared with an SVM baseline classifier using different kernel functions. In all experiments for a given number of labeled examples, half the unlabeled examples were used for cross-validation of $\alpha$ for the out-tree model and $C$ for the SVM. The remaining half of the unseen labels were used for testing. Figure 3 shows the average error rate on random folds for all tasks as the number of training examples is varied for salamander species taxonomy. Similarly, Figure 4 shows the crustaceans taxonomy. The out-tree has a statistically significant advantage over both linear and RBF SVMs when classifying data that obeys a directed tree structure.

## 10 DISCUSSION

This article described a Bayesian treatment of a latent directed out-tree connectivity on non-*iid* data-points. This led to a generative model appropriate for taxonomy and tree data as well as an interesting semi-parametric density estimator for datasets in general. The matrix tree theorem was extended to directed trees and enjoys the same efficient Bayesian inference properties as its undirected counterpart. A novel *tdid* likelihood emerges which permits the recovery of both a marginal density on nodes as well as the conditional of each node given its latent parent. The new likelihood is exchangeable and is a direct generalization of *iid* likelihood. It degenerates into *iid* when conditional dependencies between children and parents collapse. A variational Bayesian treatment is also possible by integrating over both parameters and out-tree structures jointly. Experiments with unsupervised and semisupervised learning were promising.

**Acknowledgments**

The author thanks A. Howard, C. Kemp, P. Long, J. Tenenbaum and the anonymous reviewers for their comments and for providing data. This work was supported by ONR Award N000140710507 (Mod No: 07PR04918-00) and NSF Award IIS-0347499.